\documentclass{llncs}

\usepackage{wrapfig}
\usepackage{times}
\usepackage{proof}
\usepackage{amssymb}
\usepackage{amsmath}
\usepackage{url}
\usepackage{color}
\usepackage{multicol}
\usepackage{wasysym}

\usepackage[inline]{enumitem}

\makeatletter
\renewcommand\paragraph{\@startsection{paragraph}{4}{\z@}%
                       {-2\p@ \@plus -4\p@ \@minus -4\p@}%
                       {-0.5em \@plus -0.22em \@minus -0.1em}%
                       {\normalfont\normalsize\bfseries}}
\renewcommand\section{\@startsection{section}{1}{\z@}%
                       {-8\p@ \@plus -4\p@ \@minus -4\p@}%
                       {8\p@ \@plus 4\p@ \@minus 4\p@}%
                       {\normalfont\large\bfseries\boldmath
                        \rightskip=\z@ \@plus 8em\pretolerance=10000 }}
\renewcommand\subsection{\@startsection{subsection}{2}{\z@}%
                       {-8\p@ \@plus -4\p@ \@minus -4\p@}%
                       {4\p@ \@plus 2\p@ \@minus 2\p@}%
                       {\normalfont\normalsize\bfseries}}                        

\makeatother

\newcommand{\MS}[1]{{\color{blue}{MS: #1}}}

\newcommand{\Vampire}{{\sc Vampire}}
\newcommand{\Spass}{{\sc Spass}}
\newcommand{\E}{{\sc E}}

\newcommand{\orl}{\vee}

\newcommand{\notl}{\neg}

\newcommand{\subst}[2]{#1 \mapsto #2}       
\newcommand{\Subst}[1]{\{#1\}}      

\newcommand{\eql}{\simeq}           
\newcommand{\neql}{\not\simeq}           

\newcommand{\p}{\phantom{1}}

\newcommand{\leaveout}[1]{}

\title{Selecting the Selection\thanks{
Kry\v{s}tof Hoder's contribution was carried out while at the University of Manchester. 
This work was supported by EPSRC Grant EP/K032674/1. Martin Suda and Andrei Voroknov were partially supported by ERC Starting Grant 2014 SYMCAR 639270. 
Andrei Voronkov was also partially supported by the Wallenberg Academy Fellowship 2014 - TheProSE.} }
\author{
Giles Reger\inst{1} \and 
 Martin Suda\inst{1} \and 
 Andrei Voronkov\inst{1,2,3} \and
 Kry\v{s}tof Hoder
}
\institute{
University of Manchester, Manchester, UK \and
Chalmers University of Technology, Gothenburg, Sweden \and
EasyChair
}

\begin{document}
\maketitle
\begin{abstract}
Modern saturation-based Automated Theorem Provers typically implement 
the \emph{superposition calculus} for reasoning about first-order 
logic with or without equality. 
Practical implementations of this calculus use a variety of 
\emph{literal selections}  and \emph{term orderings} to tame the 
growth of the search space and help steer proof search.
This paper introduces the notion of \emph{lookahead} selection that 
estimates (\emph{looks ahead}) the effect of 
selecting a particular literal on the number of immediate children
of the given clause and selects to minimize this value.
There is also a case made for the use of 
\emph{incomplete} selection strategies that attempt to restrict the
search space instead of satisfying some completeness criteria. 
Experimental evaluation in the \Vampire\ theorem prover shows that
both lookahead selection and incomplete selection significantly
contribute to solving hard problems unsolvable by other methods. 
\end{abstract}


\section{Introduction}
\label{sec:introduction}




This paper considers the usage of literal selection strategies 
in practical implementations of the superposition calculus (and its extensions). 
The role of literal selection in arguments for completeness have been known for a long time \cite{BachmairGanzinger-91-mpii208},
but there has been little written about their role in proof search. 
This paper is concerned with the properties of literal selections that lead to the \emph{quick} proofs i.e. those that restrict proof search in a way that can make finding a proof quickly more likely.
In fact, our disregard for completeness is strong enough to suggest \emph{incomplete} literal selections as a fruitful route to such fast proofs.
Our approach is based on the (experimental) observation that it is generally most helpful to perform inferences that lead to as few new clauses as possible. 
The main conclusion of this is a notion of \emph{lookahead selection} that selects exactly the literal that is estimated to take part in as few inferences as possible.

The setting of this work is saturation-based first-order theorem provers based on the superposition calculus.
These are predominant in the area of first-order theorem proving (see the latest iteration of the CASC competition \cite{SS06-SoCASC}).
Provers such as \E{} \cite{E}, \Spass{} \cite{spass}, and \Vampire{} \cite{KovacsVoronkov:CAV:Vampire:2013,VampireManual} 
work by saturating a clause search space with respect to an inference system (the superposition calculus)
with the aim of deriving the empty clause (witnessing unsatisfiability of the initial clause set).
Various techniques are vital to avoid explosion of the search space. 
Predominant among these is \emph{redundancy elimination} (such as subsumption) used to remove clauses. 
One can also consider methods to restrict the number of generated clauses,
this is where we will consider the role of \emph{literal selection}.
The idea is that inferences are only performed on selected literals 
and literals are selected in a way to restrict the growth of the search space.
Another effect of literal selection is to avoid obtaining the same clauses by permutations of inferences.



For the resolution calculus there is a famous result about completeness with respect to selection and term orderings \cite{BachmairGanzinger-91-mpii208} that supposes properties of the selection strategy to construct a model given a saturated set of clauses. This result carries over to superposition. As a consequence, particular selections and orderings can be used to show decidability of certain fragments of first-order logic, see e.g. \cite{DBLP:conf/kgc/BachmairGW93,DBLP:conf/lics/GanzingerN99}.
However, the requirements placed on selection by this completeness result are some times at odds with the aim of taming proof search.
This paper presents different selection strategies (including the aforementioned lookahead selection) 
that aim to effectively control proof search and argues that dropping the completeness requirements can further this goal.

The main contributions of this paper can be summarised as follows:
\begin{enumerate*}[label=\itshape\alph*\upshape)]
\item we formulate a new version of the superposition calculus 
which captures the notion of incomplete selections while being general enough
to subsume the standard presentation (Sect.~\ref{sec:selAndSuper});
\item we introduce \emph{quality selections}, an easy to implement
compositional mechanism for defining literal selections based on a notion of quality (Sect.~\ref{sec:quality}), and
\item we introduce \emph{lookahead selection} and describe how it can be efficiently implemented (Sect.~\ref{sec:lookahead}).	
\end{enumerate*}
These ideas have been realised within \Vampire{} and complemented by several selections adapted from other theorem provers (Sect.~\ref{sec:concrete}).
Our experimental evaluation (Sect.~\ref{sec:experiments}) shows 
that these new selections (incomplete and lookahead) are good at both solving the most problems overall and solving problems uniquely.

\section{Preliminaries}
\label{sec:background}



We consider the standard first-order predicate logic with
equality. Terms are of the form $f(t_1,\ldots,t_n), c$ or $x$ where
$f$ is a \emph{function symbol} of arity $n \geq 1$, $t_1,\ldots, t_n$ are terms, $c$ is a zero arity function symbol (i.e. a constant) and $x$ is a variable. 
The \emph{weight} of a term $t$ is defined as $w(t) =1$ if $t$ is a variable or a constant
and as $w(t) = 1+\sum_{i=1,\ldots,n} w(t_i)$ if $t$ is of the form
$f(t_1,\ldots,t_n)$. In other words, the weight or a term is the
number of symbols in it.
Atoms are of the form $p(t_1,\ldots,t_n), q$ or $t_1 \eql t_2$ where $p$ is a \emph{predicate symbol} of arity $n$, $t_1, \ldots, t_n$ are terms, $q$ is a zero arity predicate symbol and $\eql$ is the \emph{equality symbol}. 
The weight function naturally extends to atoms: $w(p(t_1,\ldots,t_n)) = 1+\sum_{i=1,\ldots,n} w(t_i)$, $w(t_1 \eql t_2) = w(t_1)+w(t_2)$,
and $w(q)=1$.
A literal is either an atom $A$, in which case we call it \emph{positive}, or a negation $\notl A$,
in which case we call it \emph{negative}.
We write negated equalities as $t_1 \neql t_2$. 
The weight of a literal is the weight of the corresponding atom.
We write $t[s]_p$ and $L[s]_p$ to denote that a term $s$ occurs in a term $t$ (in a literal $L$) at a position $p$.

A \emph{clause} is a disjunction of literals $L_1 \orl \ldots \orl L_n$ for $n \geq 0$.
We disregard the order of literals and treat a clause as a multiset. 
When $n=0$ we speak of the \emph{empty clause}, which is always false. 
When $n=1$ a clause is called a unit clause. 
Variables in clauses are considered to be universally quantified.
Standard methods exist to transform an arbitrary first-order formula into clausal form.

A \emph{substitution} is any expression $\theta$ of the form
$\Subst{\subst{x_1}{t_1},\ldots,\subst{x_n}{t_n}}$, where $n \geq 0$, and $E\theta$
is the expression obtained from $E$ by the
simultaneous replacement of each $x_i$ by $t_i$. By an expression
here we mean a term, an atom, a literal, or a clause. An expression is
\emph{ground} if it contains no variables.

A \emph{unifier} of two expressions 
$E_1$ and $E_2$ different from clauses is a substitution
$\theta$ such that $E_1\theta=E_2\theta$. It is known that 
if two expressions have a unifier, then they have a so-called most general unifier. Let ${\sf mgu}$ be a function returning a most general unifier of two expressions if it exists. 

A \emph{simplification ordering} (see, e.g. \cite{DBLP:books/el/RV01/DershowitzP01}) on terms is an ordering that is \emph{well-founded},
\emph{monotonic}, \emph{stable under substitutions} and has the \emph{subterm property}. Such an ordering captures a notion of \emph{simplicity} i.e. $t_1 \prec t_2$ implies that $t_1$ is in some way simpler than $t_2$. \Vampire{} uses the Knuth-Bendix ordering \cite{KnuthDE:simwpu}.
Such term orderings are usually total on ground terms and partial on non-ground ones.
%
There is a simple extension of the term ordering to literals, the details of which are not relevant here.


\leaveout{

\paragraph{Saturation-based Proof Search}
Saturation-based theorem provers \emph{saturate} a set of clauses $S$ with respect to an inference system $\mathbb{I}$ e.g. they compute a set of clauses $S'$ by applying rules in $\mathbb{I}$ to clauses in $S$ until no new clauses are generated. 
If the empty clause is generated then $S$ is unsatisfiable.
Calculi such as resolution and superposition have conditions that ensure \emph{completeness}, which means that a saturated set $S$ is satisfiable if it does not contain the empty clause as an element.
As first-order logic is only semi-decidable, it is not necessarily the case that $S$ has a finite saturation, and even if it does it may be unachievable in practice using the available resources. 
Therefore, much effort in saturation-based first-order theorem proving involves 
controlling proof search to make finding the empty clause more likely (within reasonable resource bounds). 
One important notion is that of \emph{redundancy}, being able to remove clauses from the search space that are not required. 
Another important notion, the subject of this paper, are literal selections that place restrictions on the inferences that can be performed. 
Both notions come with additional requirements for completeness.
}



\section{The Superposition Calculus and Literal Selection}
\label{sec:selAndSuper}

The superposition calculus as implemented in modern theorem provers 
usually derives from the work of Bachmair and Ganzinger \cite{BachmairGanzinger-91-mpii208}
(see also \cite{BachmairGanzinger:HandbookAR:resolution:2001,NieuwenhuisRubio:HandbookAR:paramodulation:2001}).
There, the inference rules of the calculus come equipped
with a list of side conditions which restrict the applicability
of each rule. The rules are sound already in their pure form,
but the additional side conditions are essential in practice 
as they prevent the clause search space from growing too fast.
At the same time, it is guaranteed that the calculus remains
refutationally complete, i.e.~able to derive the empty clause
from every unsatisfiable input clause set.

Here we are particularly interested in side conditions 
concerning individual literals within a clause on which an inference should be performed.
The formulation by Bachmair and Ganzinger derives these conditions
from a simplification ordering $\prec$ on terms and its extension to literals,
and from a so called \emph{selection function} $S$ which
assigns to each clause $C$ a possibly empty multiset $S(C)$
of negative literals in $C$, which are called selected.
The ordering and the selection function should be understood as parameters of the calculus.

The calculus is designed in such a way that an inference
on a positive literal $L$ within a clause $C$ must only
be performed when $L$ is a maximal literal in $C$
(i.e. there is no literal $L'$ in $C$ such that $L \prec L'$)
and there is no selected literal in $C$. Complementarily,
an inference on a negative literal $L$ within a clause $C$ must only
be performed when $L$ is a maximal literal in $C$ and there is no selected literal in $C$
or $L$ is selected in $C$. 
Such conditions are shown to be compatible with completeness.

In this paper, we take a different perspective on literal selection.
We propose the notion of a \emph{literal selection strategy}, or
\emph{literal selection} for short,
which is a procedure 
that assigns to a non-empty clause $C$ a non-empty
multiset of its literals. We avoid the use of the word ``function'' on
purpose, since it is not guaranteed that we select the same multiset even
if the same clause occurs in a search space again after being deleted.
In addition, we do not want the selection to depend just on the clause itself,
but potentially also on a broader context including the current state of the search space.

We formulate the inference rules 
of superposition such that an inference on a literal within 
a clause is only performed when that literal is selected.
This is evidently a simpler concept, which primarily decouples
literal selection from completeness considerations as it also allows
incomplete literal selection.
At the same time, however, it is general enough so that completeness 
can be easily taken into account when a particular selection strategy is designed. 


\begin{figure}[tb]
\[
\begin{array}{l}
\begin{array}{lll}
\mathbf{Resolution} &\quad\quad\quad& \mathbf{Factoring} \\
\\
\infer[,]{(C_1 \vee C_2)\theta}{\underline{A} \vee C_1 & \underline{\neg A'} \vee C_2} &&
\infer[,]{(A \vee C)\theta}{\underline{A} \vee A' \vee C} \\
\end{array}\\
\\
\textit{where, for both inferences, } \theta = {\sf mgu}(A,A') \textit{ and } A\textit{ is not an equality literal} \\
\\
\mathbf{Superposition}\\
\\
\begin{array}{lll}
\infer[\quad\textit{or}\quad]{(L[r]_p \vee C_1 \vee C_2)\theta}{\underline{l \eql r} \vee C_1 & \underline{L[s]_p} \vee C_2} &&
\infer[,]{(t[r]_p \otimes t' \vee C_1 \vee C_2)\theta}{\underline{l \eql r} \vee C_1 & \underline{t[s]_p \otimes t'} \vee C_2}
\\
\end{array}\\
\\
\textit{where}~\theta = {\sf mgu}(l,s) ~\textit{and}~ r\theta \not \succeq l\theta ~\textit{and, for the left rule}~L[s] ~\textit{is not an equality literal,}\\ \textit{and for the right rule } \otimes \textit{ stands either for }  \eql  \textit{ or }  \neql  \textit{ and } t'\theta \not \succeq t[s]\theta 
\\
\\
\begin{array}{lll}
\mathbf{Equality Resolution} &\quad\quad\quad& \mathbf{Equality Factoring} \\
\\
\infer[,]{C\theta}{\underline{s \neql t} \vee C} &&
\infer[,]{(t \neql  t' \vee s' \eql t' \vee C)\theta}{\underline{s \eql t} \vee s'  \eql  t' \vee C} \\
\textit{where } \theta = {\sf mgu}(s,t) && \textit{where } \theta = {\sf mgu}(s,s'),~ t\theta \not \succeq s\theta, \textit{ and } t'\theta \not \succeq s'\theta\\
\end{array}\\
\end{array}
\]
\caption{The rules of the superposition and resolution calculus.}
\label{fig:superposition:calculus}
\end{figure}

\paragraph{The Calculus.}
Our formulation of the superposition and resolution calculus with 
literal selection is presented in Fig.~\ref{fig:superposition:calculus}.
It consists of the resolution and factoring rules for dealing with non-equational literals
and the superposition, equality resolution and equality factoring rules 
for equality reasoning. Although resolution and factoring can be simulated
by the remaining rules provided non-equational atoms are encoded 
in a suitable way,
we prefer to present them separately, because they also have separate implementations
in \Vampire{} for efficiency reasons.

The calculus in Fig.~\ref{fig:superposition:calculus} is parametrised by a simplification ordering $\prec$ 
and a literal selection strategy, which we indicate here (and also in the rest of the paper) by underlining. 
In more detail, literals underlined in a clause  must be selected by the strategy. Literals without underlying may be selected as well. 
Generally, inferences are only performed between selected literals with the exception of the two factoring rules.
There only one atom needs to be selected and factorings are performed with other unifiable atoms.

We remark that further restrictions on the calculus can be added on top of those mentioned in Fig.~\ref{fig:superposition:calculus}.
In particular, if literal selection captures the maximality condition of a specific literal in a premise, 
this maximality may be required to also hold for the instance of the premise
obtained by applying the {\sf mgu} $\theta$. We observed that these additional
restrictions did not affect the practical performance of our prover
in a significant way and for simplicity kept them disabled during our
experiments.

We also note that the calculi based on the standard notion of
selection function can be captured by our calculi -- all we have to do
is to select all maximal literals in clauses with no literals
selected by the function.



\paragraph{Selection and Completeness.}
We now reformulate the previously mentioned side conditions on literals 
which are required by the completeness proof of Bachmair and Ganzinger \cite{BachmairGanzinger-91-mpii208}
in terms of literal selection strategies.
In the rest of the paper we refer to strategies satisfying the following \emph{completeness condition} as \emph{complete selections}:
\begin{equation}
\text{Select either a negative literal or all maximal literals with respect to $\prec$.} \label{cond:complete}
\end{equation}
Although selections which violate condition (\ref{cond:complete})  cannot be used 
for showing satisfiability of a clause set by saturation, our experimental results will demonstrate
that incomplete selections are invaluable ingredients for solving many problems.

As an example of what can happen if condition (\ref{cond:complete}) is violated,
consider the following unsatisfiable set of clauses where \emph{all} selected literals are underlined.
\[
p \orl \underline{q} \quad\quad
\underline{p} \orl \notl q \quad\quad
\underline{\notl p} \orl q \quad\quad
\notl p \orl \underline{\notl q} 
\]
Note that this set is clearly unsatisfiable as one can easily derive $p$ and $\neg p$ and then the empty clause. However, using the given selection it is only possible to derive tautologies. The selection strategy does not fulfill the above requirements as either $p \succ q$ and $p$ must be selected in $p \orl q$, or $q \succ p$ and $\neg q$ must be selected in $p \orl \notl q$.


\section{Quality Selections}
\label{sec:quality}

Vampire implements various literal selections in a uniform way, using 
preorders on literals, which try to reflect certain notions of quality. We convert such a preorder to a linear order by breaking ties in an arbitrary but fixed way. This order on literals (a \emph{quality order}) induces two selections, one incomplete and one
complete. Essentially, the incomplete one simply selects the literal greatest in this order and the complete one modifies the incomplete literal selection where the latter violates the sufficient conditions
for completeness. We call the resulting class of selections \emph{quality selections}. We believe that this is a new way of defining literal selections that has not been reported in the literature or observed in other systems before.

The preorders we use capture various notions of quality the literals we want to select should have. 
Let us now discuss what it is that we want to achieve from selection.
The perfect selection strategy contains an oracle that knows the exact inferences necessary to derive the empty clause in the shortest possible time. 
Without such an oracle we can employ heuristics to suggest those inferences that are more desirable. 

There is a general insight that a slowly growing search space is superior 
to a faster growing one, provided completeness is not compromised too much. 
It should be evident that a search space that grows too quickly will soon become unmanageable,
 reducing the likelihood that a proof is found. This has been repeatedly observed in practice. 
This insight holds despite the fact that the shortest proofs for some formulas may theoretically become much longer in the restricted (slowly growing) setting. 
\emph{Therefore, the aim of a selection strategy in our setting is to generate the fewest new clauses.}

\subsection{Quality Orderings}

\newcommand{\qo}{\ensuremath{\vartriangleright}}

Let us consider several preorders \qo{} on literals that capture notions of preference for selection i.e. $l_1 \qo{} l_2$ means we should prefer selecting $l_1$ to $l_2$. If they are equally preferable, that is $l_1 \qo{} l_2$ and $l_2 \qo{} l_1$, we will write $l_1 \equiv l_2$. 
We are interested in preorders that prefer literals having as few children as possible, this means decreasing the likelihood that we can apply the inferences in Fig.~\ref{fig:superposition:calculus}.


\paragraph{Unifiability.} Firstly we note that all inferences require the selected literal (or one of its subterms) to unify with something in another clause. Therefore, we prefer literals that are potentially unifiable with fewer literals in the search space.

To this end, we first note that a heavy literal is likely to have a complex structure containing multiple function symbols. It is therefore unlikely that two heavy literals will be unifiable. This observation is slightly superficial because, for example, a literal $p(x_1,\ldots,x_n)$ for large $n$ has a large weight but unifies with all negative literals containing $p$. Let $l_1 \qo_{weight} l_2$ if the weight of $l_1$ is greater than the weight of $l_2$.
%

Next, we note that the fewer variables a literal contains the less chance it has to unify with other literals 
e.g. $p(f(x),y)$ will unify with every literal that $p(f(a),y)$ will unify with,
 and potentially many more. Let $l_1 \qo_{vars} l_2$ if $l_1$ has fewer variables than $l_2$.
 %

However, we can observe that not all variables are equal, the literal $p(x)$ will unify with more than $p(f(f(x)))$. As a simple measure of this we can consider only variables that occur at the top-level i.e. immediately below a predicate symbol. Let $l_1 \qo_{top} l_2$ if $l_1$ has fewer top-level variables than $l_2$.
%
%
Similarly, $p(f(x),f(y))$ will unify with more than $p(f(x),f(x))$ as the repetition of $x$ constrains the unifier. To capture this effect we can prefer literals with fewer \emph{distinct} variables. Let $l_1 \qo_{dvar} l_2$ if $l_1$ has fewer distinct variables than $l_2$.
%

\paragraph{Equality and Polarity.}
We can observe from the inference rules in Fig.~\ref{fig:superposition:calculus} that positive equality is required for superposition, which can be a prolific inference as it can rewrite inside a clause many times. Therefore, we should prefer not to select positive equality where possible. Let $L \qo_{nposeq} s \eql t$, where $L$ is a non-equality literal, and $s \neql t \qo_{nposeq} s' \eql t'$.

%
In a similar spirit, we observe that negative equality otherwise only appears in Equality Resolution
which is in general a non-problematic inference 
as it is performed on a single clause and decreases the number of its literals.
Therefore, in certain cases we should prefer negative equalities. Let $s \neql t \qo_{neq} L$ where $L$ is a non-equality literal.
%

Finally, for non-equality literals it is best to default to selecting a single polarity as literals with the same polarity cannot resolve.
Furthermore, selecting negative literals seems to be preferable as it keeps the corresponding selection strategy 
from compromising the completeness condition. We let $\neg A \qo_{neg} A'$.




\subsection{Quality-Based Selections}

We want to compose different notions of quality so that we can break ties when the first notion is too coarse to distinguish literals. 
We define the composition of two preorders ${\qo}_a$ and ${\qo}_b$, denoted by $\qo_a \circ \qo_b$, by
$l_1 ~(\qo_a \circ \qo_b)~ l_2$ if and only 
$l_1 \qo_a l_2$, or $l_1 \equiv_a l_2$ and $l_1 \qo_b l_2$. Evidently, a composition of two preorders is also a preorder.

Given a preorder $\qo$ we define a selection strategy $\pi_{\qo}$
that selects the greatest (highest quality) literal with respect to $\qo$
breaking ties arbitrarily, but in a deterministic fashion. We call such strategies \emph{quality selections.}



\subsection{Completing the Selection} \label{sec:quality:complete}

Quality selections are not necessarily complete i.e. 
they do not satisfy the completeness condition (\ref{cond:complete}) introduced in Sect.~\ref{sec:selAndSuper}.
It is our hypothesis that these incomplete selection strategies are practically useful.
However, there are cases where complete selection is desirable. One obvious example is where we are attempting to establish satisfiability.\footnote{%
It should be noted that \Vampire\ always knows when it is incomplete and therefore returns Unknown when
obtaining a saturated set with the help of an incomplete strategy. }

Given a quality selection $\pi_{\qo}$, it is possible to also define a \emph{complete} selection strategy using the following steps. Let $N$ initially be the set of all literals in a clause and $M$ be the subset of $N$ consisting of all its literals maximal in the simplification ordering.
\begin{enumerate}
	\item {\bf If} $\pi_{\qo}(N)$ is negative {\bf then} select $\pi_{\qo}(N)$
	\item {\bf If} $\pi_{\qo}(N) \in M$ and all literals in $M$ are positive {\bf then} select $M$
	\item {\bf If} $M$ contains a negative literal {\bf then} set $N$ to be the set of all negative literals in $M$ and {\bf goto} 1
	\item Remove $\pi_{\qo}(N)$ from $N$ and {\bf goto} 1
\end{enumerate}
This attempts to, where possible, select a single negative literal that is maximal with respect to the quality ordering. The hypothesis being that it is always preferable to select a single negative literal rather than several maximal ones.


\section{Lookahead Selection}
\label{sec:lookahead}

In this section we introduce a general notion of \emph{lookahead selection} and describe an efficient implementation of the idea. Our discussion in the previous section suggested that we try to find preorders that potentially minimize the number of children of a selected literal. Essentially, lookahead selection tries to select literals that result in the smallest number of children. Note that this idea requires a considerable change in the design and implementation, because the number of children depends on the current state of the search space rather than on measures using only the clause we are dealing with.

\subsection{Given-Clause Algorithms and Term Indexing}


Before we can describe lookahead selection we 
give some context about how \Vampire\ and other modern provers implement saturation-based proof search. 


\Vampire\ implements a given-clause algorithm that maintains a set of \emph{passive} and a set of \emph{active} clauses and executes a loop where (i) a given clause is chosen from the passive set and added to the active set, (ii) all (generating) inferences between the given clause and clauses in active are performed, and (iii) new clauses are considered for forward and backward simplifications and added to passive if they survive. The details of (iii) are not highly relevant to this discussion, but are very important for effective proof search. 

Generating inferences are implemented using \emph{term indexing} techniques (see e.g. \cite{TermIndexing}) that index a set of clauses (the active clauses in this case) and can be queried for clauses containing subexpressions that match or unify with a given expression.

We can view a term index ${\cal T}$ for an inference rule as a map that takes a clause $\underline{l} \orl D$ with a selected literal $l$ and returns a list of \emph{candidate clauses}, which is a set containing all clauses that can have this inference against $\underline{l} \orl D$. 
%
%
\Vampire\ maintains two term indexes for superposition and a separate one for binary resolution. Term indexes are not required for factoring or equality resolution as these are performed on a single clause.

\subsection{General Idea Behind Lookahead Selection}

The idea of lookahead selection is that we directly estimate for each literal $l$ in $C$
how many children the clause $C$ would have when selecting $l$ and applying inferences on $l$ against active clauses.

Ideally we would have access to a function ${\sf children}(C, l)$ that would return the number of children of clause $C$ resulting from inferences with active clauses, given that the literal $l$ was selected in $C$. We discuss how we practically estimate such a value below.

Given this value we can define a preorder\footnote{Note that this is not a preorder in the same sense as before as it requires the context of a clause and active clause set. In other words, this preorder is a relation that changes during the proof search process.} that minimises the number of children:
\[
l_1 \qo_{lmin} l_2 ~~\mathit{ iff }~~ {\sf children}(C,l_1) < {\sf children}(C,l_2) 
\]
This is based on our previous assertion that we want to produce as few children as possible. But now we have an effective way of steering this property we can also consider the opposite i.e. introduce a quality ordering that maximises the number of children:
\[
l_1 \qo_{lmax} l_2 ~~\mathit{ iff }~~ {\sf children}(C,l_1) > {\sf children}(C,l_2) 
\]
Our hypothesis is that a selection strategy based on this second ordering will perform poorly, as the search space would grow too quickly.

\subsection{Completing the Selection... Differently}

In Sect.~\ref{sec:quality:complete} selection strategies were made complete by searching for the best negative literal where possible. The same approach is taken for selection strategies based on lookahead selection but because it is now relatively much more expensive to compare literals it is best to decide on the literals to compare beforehand.

Firstly, if there are no negative literals all maximal literals must be selected and no lookahead selection is performed. Otherwise, selection is performed on all negative literals and a single maximal positive literal (if there is only one). This ignores the complex case where the combination of all maximal literals would lead to fewer children than the best negative literal.

\subsection{Efficiently Estimating Children}

To efficiently estimate the number of children that would arise from selecting a particular literal in a clause we make use of the term indexing structures. 

Let ${\cal T}_1, \ldots, {\cal T}_n$ be a set of term indexes capturing the current active clause set. An estimate for ${\sf children}(C, l)$ can then be given by:
\[
{\textsf{estimate}}(l) = \Sigma_{i=1}^{n} |{\cal T}_i[l]| .
\]
This is an overestimate as the term indexes do not check side-conditions related to orderings after substitution. For example, if we apply a superposition from $l \eql r$ with $\theta = mgu(l,r)$ and we have $r \not\succeq l$, $r\theta \succeq l\theta$, the index will select $l \eql r \orl C_1$ as a candidate clause but the rule does not apply. In addition, the number of children is not the same as the number of children that survive retention tests (those neither deleted nor simplified away). However, applying all rules and simplifying children for every literal can be very time-consuming, so we use an easier-to-compute approximation instead.

%
It is possible to extend the estimate to include inferences that do not rely on indexes. We have done this for equality resolution but not factoring, due to the comparative effort required. 
In general, our initial hypothesis was that selection should be a cheap operation and so
it is best to perform as few additional checks as possible.

In \Vampire\ term indexes return iterators over clauses. This allows us to compute ${\sf estimate}$ in a fail-fast fashion where we search all literals at once and terminate as soon as the estimate for a single literal is finished. This assumes we are minimising (i.e. computing maximal literals with respect to $\qo_{lmin}$), otherwise we must exhaust the iterators of all but one of the literals. 

Of course, as selecting literals in this way now depends on 
the active clauses it is desirable to do selection as late as possible
to maximise accuracy of the estimate. 
Therefore, \Vampire\ performs literal selection at the point 
when it chooses a clause from the passive set for activation.

Note that the technique described here can be extended to any setting that uses indexes for generating inferences.

\section{Concrete Literal Selection Strategies}
\label{sec:concrete}

In this section we briefly describe concrete literal selection strategies. To have a more general view of selections, we also implemented some selections found in other systems. Of course, when considering selections adapted from other systems we cannot draw conclusions about their utility in the original system as the general implementation is different. But it is useful to compare the general ideas. Strategies have been given numbers to identify them that is based on an original numbering in \Vampire, these numbers are used in the next section.


\subsection{Vampire}
\label{sec:vampire:selections}

We give a brief overview of the selection strategies currently implemented in \Vampire.

\paragraph{Total Selection.} The most trivial literal selection strategy is to select everything. This corresponds to the calculus without a notion of selection and is obviously complete. This is referred to by number 0.

\paragraph{Maximal Selection.} \Vampire{}'s version of maximal selection either
selects one maximal negative literal, if one of the maximal literals is negative,
or all maximal literals, in which case they will all be positive. This is referred to by number 1.

\paragraph{Quality Selections.} \Vampire\ uses four quality selections obtained by combining preorders defined in the previous section as follows:
\[
\begin{array}{lll}
\qo_2 &=& \qo_{weight} \\
\qo_3 &=& \qo_{noposeq} \circ \qo_{top} \circ \qo_{dvar} \\
\qo_4 &=& \qo_{noposeq} \circ \qo_{top} \circ \qo_{var} \circ \qo_{weight} \\
\qo_{10} &=& \qo_{neq} \circ \qo_{weight} \circ \qo_{neg} \\
\end{array}
\]
\Vampire\ uses both the incomplete versions of the selection strategies, which it numbers 1002, 1003, 1004 and 1010, and the complete versions, which it numbers 2, 3, 4 and 10. We note that not all combinations of the preorders discussed in Sect.~\ref{sec:quality} are used. As may be suggested by the numbering, previous experimentation introduced and removed various combinations thereof, leaving the current four. 

\paragraph{Lookahead Selection.} \Vampire\ uses two lookahead selections based on preorders defined as follows
\[
\begin{array}{lll}
\qo_{11} &=& \qo_{lmin} \circ \qo_3 \\
\qo_{12} &=& \qo_{lmax} \circ \qo_3 \\
\end{array}
\]
The incomplete versions of the associated strategies are numbered 1011 and 1012 whilst the complete versions are numbered 11 and 12.


\subsection{SPASS Inspired}
\label{sec:spass:selections}

We consider three literal selection strategies adapted from \Spass{} (as found in the prover's source code)\footnote{
\Spass{} also has ``select from list'', which requires the user to specify predicates that will
be preferred for selection. We did not implement this for the obvious reason.}:
\begin{itemize}
\item{\bf Selection off (20)} selects all the maximal literals. From the perspective of the original Bachmair and Ganzinger theory nothing is selected, but in our setting this effectively amounts to selecting all the maximal literals.

\item {\bf Selection always (22)} selects a negative literal with maximal weight, if there is one. 
Otherwise it selects all the maximal ones.

\item {\bf If several maximal (21)} selects a unique maximal, if there is one.
Otherwise it selects a negative literal with maximal weight, if there is one.
And otherwise it selects all the maximal ones.
\end{itemize}

\subsection{E Prover Inspired}
\label{sec:e:selections}

We consider the following five literal selection strategies adapted from \E{} 
(as mentioned in the prover's manual \cite{emanual}):
\begin{itemize}	
\item
	{\bf SelectNegativeLiterals (30)} selects all negative literals, if there are any.
	Otherwise it selects all the maximal ones.
	
\item
	{\bf SelectPureVarNegLiterals (31)} selects a negative equality between variables,
	if there is one. Otherwise it selects all the maximal literals.

\item
	{\bf SelectSmallestNegLit (32)} selects a negative literal with minimal weight, if there is one.
	Otherwise it selects all the maximal literals.	

\item
	{\bf SelectDiffNegLit (33)} selects a negative literal which maximises the difference
	between the weight of the left-hand side and the right-hand side,\footnote{
	In E, all literals are represented as equalities. A non-equational atom $p(t)$
	is represented as $p(t)=\top$, where $\top$ is a special constant \emph{true}.
	Thus it makes sense to talk about left-hand and right-hand side
	of a literal even in the non-equational case.} if there is a negative literal at all.
	Otherwise it selects all the maximal literals.
	
\item
	{\bf SelectGroundNegLit (34)} selects a negative ground literal for which the weight
	difference between the left-hand side and the right-hand side terms is maximal,
	if there is a negative literal at all.
	Otherwise it selects all the maximal literals.	


\item
	{\bf ``SelectOptimalLit'' (35)} selects as (34) if there is a ground negative literal
	and as (33) otherwise.
	
		
\end{itemize}

It should be noted that our adaptations of \E{}'s selections are only approximate,
because E{} uses a different notion of term weight than \Vampire{}, 
defining constants and function symbols to have basic weight 2 and variables to have weight 1.
Also we do not consider \E{}'s {\bf NoSelection} strategy separately as it is the same as \Spass{}'s {\bf Selection off} 
and \E{}'s {\bf SelectLargestNegLit} strategy as it is the same as \Spass{}'s {\bf Selection always}
(modulo the notion of term weight).



\section{Experimental Evaluation}
\label{sec:experiments}

Here we report on our experiments with selection strategies 
using the theorem prover \Vampire{}. Our aim is to look for strategies
which help to solve many problems, but also for strategies which solve
problems other strategies cannot solve. This is because 
we are ultimately interested in constructing a portfolio combining 
several strategies which solve as many problems as possible 
within a reasonably short amount of time.



\paragraph{Experimental Setup.} 
For our experiments we took all the problems from the TPTP \cite{TPTP} library version 6.3.0 which are in the FOF or CNF format,
excluding only unit equality problems (for which literal selection does not play any role) 
and problems of rating 0.0 (which are trivial to solve). This resulted in a collection of $11\,107$ problems.\footnote{
A list of the selected problems, the executable of our prover as well as the results 
of the experiment are available at \url{http://www.cs.man.ac.uk/~sudam/selections.zip}.} 

We ran \Vampire{} on these problems with saturation algorithm set to \emph{discount}
and \emph{age-weight ratio} to $1:5$ (cf. \cite{KovacsVoronkov:CAV:Vampire:2013,VampireManual}),
otherwise keeping the default settings 
and varying the choice of literal selection.
By default, \Vampire{} employs the AVATAR architecture 
to perform clause splitting \cite{avatar,DBLP:conf/cade/RegerSV15}.
AVATAR was also enabled in our experiments. 

The time limit was set to 10 seconds for a strategy-problem pair.
This should be sufficient for obtaining a realistic picture of relative 
usefulness of each selection strategy, given the empirical observation
pertaining to first-order theorem proving in general, 
that a strategy usually solves a problem very fast if at all.
The experiments were run on the StarExec cluster \cite{starexec},
whose nodes are equipped with Intel Xeon 2.4GHz processors. 
Experiments used Vampire's default memory limit of 3GB.

\paragraph{Result Overview.}


In total, we tested 23 selection strategies i.e. those summarised in Sect.~\ref{sec:concrete}.
With AVATAR, \Vampire{} never considers clauses with ground literals for selection,
therefore selection 34 behaves the same as 20 and 35 the same as 33.
Consequently, results for 34 and 35 are left out from initial discussions, 
but will be discussed later when we consider what happens when AVATAR is not used.

Out of our problem set, $5\,908$ problems were solved by at least one strategy.\footnote{
And $1\,952$ problems were solved by every strategy.}
This includes 31 problems of TPTP rating 1.0. 
Out of the solved problems, $5\,621$ are unsatisfiable  
and $287$ satisfiable. 
Because we are mainly focusing on theorem proving, i.e. showing unsatisfiability,
we will first restrict our attention to the unsatisfiable problems.

\paragraph{Ranking the Selections.}

Table~\ref{tab:initial_view} (left) shows the performance of the individual selection strategies.
We report the number of problems solved by each strategy 
(which determines the order in the table),
the percentage with respect to the above reported overall total of problems solved,
the number of problems solved by only the given strategy (unique),
and an indicator we named \emph{u-score}. 
U-score is a more refined version of the number of uniquely solved problems.
It accumulates for each problem solved by a strategy the 
reciprocal of the number of strategies which solve that problem. This means that each uniquely
solved problem contributes $1.0$, each problem solved also by one other strategy adds $0.5$, etc.
It also means that the sum of u-scores in the whole table equals the number of problems solved in total.

\begin{table}[tb]
\caption{Left: performance of the individual selection strategies. 
Right: statistics collected from the runs: \#child is the average 
number of children of an activated clause, \%incomp is the average 
percentage of the cases when an incomplete selection violates the completeness condition. 
The values marked `s.o.' (solved only) are collected only from runs which solved a problem,
the values marked `all' are collected from all runs.}

\label{tab:initial_view}
\centering
\begin{tabular}{rcrrrc|ccc}
selection & \#solved & \%total & \#unique & u-score &\p{}&\p{}& \#child (s.o./all) & \%incomp. (s.o./all) \\
\hline
1011 & 4718 & 83.9 & 156 & 563.6 &&& \p{}4.2 / \p{}9.9 & \p{}3.3 / 4.5 \\
1010 & 4461 & 79.3 & 31 & 384.1 &&& \p{}9.4 / 14.6 & \p{}2.1 / 2.5 \\ 
11 & 4333 & 77.0 & 26 & 354.7 &&& \p{}6.5 / 13.6 \\ 
1002 & 4327 & 76.9 & 62 & 396.1 &&& \p{}8.7 / 15.4 & \p{}9.7 / 7.6 \\
10 & 4226 & 75.1 & 8 & 283.3 &&& \p{}9.9 / 14.5 \\
21 & 4113 & 73.1 & 6 & 274.2 &&& 10.7 / 13.8 \\
2 & 4081 & 72.6 & 1 & 261.0 &&& 10.3 / 14.9 \\ 
1004 & 4009 & 71.3 & 8 & 276.2 &&& \p{}6.3 / 14.1 & 19.5 / 7.3 \\
4 & 3987 & 70.9 & 2 & 247.2 &&& \p{}7.8 / 13.7 \\
3 & 3929 & 69.8 & 1 & 235.5 &&& \p{}8.7 / 13.8 \\
1003 & 3907 & 69.5 & 6 & 258.2 &&& \p{}6.5 / 14.7 & 22.6 / 8.6 \\
33 & 3889 & 69.1 & 1 & 239.2 &&& \p{}7.1 / 18.3 \\
22 & 3885 & 69.1 & 0 & 236.2 &&& \p{}7.0 / 18.4 \\
1 & 3778 & 67.2 & 6 & 227.9 &&& \p{}9.4 / 19.9 \\
31 & 3702 & 65.8 & 0 & 218.2 &&& 13.4 / 23.1 \\
20 & 3682 & 65.5 & 0 & 217.1 &&& 13.3 / 23.2 \\
30 & 3559 & 63.3 & 3 & 204.9 &&& 16.6 / 28.8 \\
32 & 3538 & 62.9 & 5 & 209.8 &&& \p{}6.3 / 19.9 \\
0 & 3362 & 59.8 & 8 & 203.1 &&& 35.8 / 48.7 \\
12 & 3308 & 58.8 & 3 & 183.4 &&& 14.0 / 24.5 \\
1012 & 2532 & 45.0 & 5 & 146.1 &&& 13.9 / 30.8 & \p{}7.6 / 5.8 \\
\end{tabular}
\end{table}

By looking at Table~\ref{tab:initial_view} we observe
that 1011, the incomplete version of the lookahead selection,
is a clear winner both with respect to the number of solved
problems and the number of uniques. 
It solves more than 80\% of problems solvable by at least one strategy
and accumulates by far the highest u-score.
Other very successful 
selections are the incomplete 1010 and 1002, and 11, 
the complete version of lookahead.

Inverted lookahead in the incomplete (1012) and complete (12) version  
end up last in the table, which can be seen as a confirmation of our hypothesis from Sect.~\ref{sec:lookahead}.
Similarly the experimental selection 0, which selects all the literals in a clause,
and the selection 32 adapted from E, which selects the \emph{smallest} negative literal,
inverting the intuition that large (with large weight) literals should be selected,
end up at the end of the table.\footnote{
The selection strategy selecting the largest negative literal has number 22.}
Interestingly, however, to each of these ``controversial'' selections 
we can attribute several uniquely solved problems.

Table~\ref{tab:initial_view} also shows that, with the exception of selection 3 (and 12),
the incomplete version of a selection always solves more problems than the complete one.

\paragraph{Additional Statistics.}
Table~\ref{tab:initial_view} (right) displays for each selection
two interesting averages obtained across the runs.
The first is the average number of children of an activated clause
and the results confirm that the lookahead selections (1011 and 11),
in accord with their design, achieve the smallest value for this metric.
This further confirms our hypothesis that preferring to generate as 
few children as possible leads to successful strategies.
The second is the average number of times an incomplete strategy 
selects in such a way as to violate the completeness condition. 
We can see that there is lot of variance between the selections
in this regard and that the ``most complete'' incomplete selection
is the second best selection 1010.

\paragraph{Time spent on performing selection.} 
As we might expect, lookahead selections are far more expensive to compute. On average, performing quality selection consumes roughly 0.1\% of the time spent on proof search, with other non-lookahead selections taking similar times, whereas complete and incomplete lookahead selection consumes roughly 1.74\% and 4.27\% respectively. 
These numbers are taken from all proof attempts, not just successful ones. Incomplete lookahed selection is more expensive than its complete counterpart as the latter is not performed when there are no negative literals.
%
%
The previously observed success of lookahead selection confirms that
the extra time
spent on selecting is more than well spent. 

\paragraph{The Effect of Turning Splitting off.} \label{sec:splitting:off}

The previous results were obtained running \Vampire\ with splitting turned on. In order to establish how much the standing of the individual selections
depends on running within the context of the AVAVAR architecture,
we ran a separate experiment with the same strategies but turning AVATAR off. 
Arguably, these results are more relevant to implementations that do not incorporate the effective AVATAR approach.

In total, the strategies without AVATAR solved $5\,563$ problems ($5\,356$ unsatisfiable, $207$ satisfiable).
The number of problems solved by all strategies was $1\,748$. 
Table~\ref{tab:initial_view_sploff} presents a view analogous to Table~\ref{tab:initial_view}
for these strategies. Strategies 34 and 35 are now relevant (see experimental setup above). 
 %
Notably, selection 1011 has dropped to the second place in the overall ranking 
(after 1010). 
However, the incomplete lookahead still accumulated the highest 
u-score as a standalone strategy and we see the same general trend that incomplete versions of strategies outperform their complete counterpart (again with the exception of 3 and 12).



\begin{table}[tb]
\caption{Performance of the individual selection strategies (left) and statistics collected from the runs (right) for runs with AVATAR turned off. Columns analogous to those described in Table~\ref{tab:initial_view}. }
\label{tab:initial_view_sploff}
\centering
\begin{tabular}{rcrrrc|ccc}
selection & \#solved & \%total & \#unique & u-score &\p{}&\p{}& \#child (s.o./all) & \%incomp. (s.o./all) \\
\hline
1010 & 4289 & 80.0 & 64 & 379.8 &&& \p{}9.3 / 17.0 & \p{}9.0 / \p{}9.4 \\ 
1011 & 4255 & 79.4 & 104 & 412.7 &&& \p{}8.5 / 15.0 & \p{}6.5 / \p{}8.3 \\
1002 & 4207 & 78.5 & 45 & 356.2 &&& \p{}7.5 / 18.5 & 17.6 / \p{}8.6 \\ 
11 & 4121 & 76.9 & 25 & 292.9 &&& 12.1 / 25.7 \\ 
10 & 4116 & 76.8 & 9 & 251.7 &&& 13.1 / 21.2 \\ 
2 & 4063 & 75.8 & 0 & 235.7 &&& 16.5 / 23.5 \\ 
21 & 4055 & 75.7 & 4 & 244.1 &&& 16.3 / 23.6 \\ 
22 & 3896 & 72.7 & 0 & 218.0 &&& \p{}8.8 / 30.3 \\ 
33 & 3895 & 72.7 & 1 & 218.0 &&& \p{}9.0 / 30.1 \\ 
4 & 3892 & 72.6 & 3 & 216.5 &&& \p{}9.6 / 19.8 \\ 
35 & 3858 & 72.0 & 1 & 211.7 &&& \p{}9.0 / 30.2 \\ 
1004 & 3810 & 71.1 & 8 & 228.2 &&& \p{}8.6 / 20.4 & 23.8 / 10.5 \\ 
3 & 3755 & 70.1 & 3 & 205.4 &&& 12.0 / 20.8 \\ 
1 & 3744 & 69.9 & 2 & 207.3 &&& 13.1 / 31.5 \\ 
30 & 3731 & 69.6 & 11 & 220.0 &&& \p{}8.9 / 33.8 \\ 
1003 & 3654 & 68.2 & 2 & 211.2 &&& \p{}8.2 / 22.8 & 25.7 / 11.1 \\ 
31 & 3517 & 65.6 & 0 & 184.9 &&& 22.4 / 33.2 \\ 
34 & 3491 & 65.1 & 1 & 183.0 &&& 21.5 / 31.9 \\
32 & 3482 & 65.0 & 2 & 188.5 &&& \p{}7.8 / 31.9 \\ 
20 & 3479 & 64.9 & 0 & 182.2 &&& 21.7 / 33.3 \\ 
12 & 3313 & 61.8 & 6 & 173.8 &&& 25.0 / 33.9  \\ 
0 & 3279 & 61.2 & 24 & 206.4 &&& 59.2 / 83.1 \\ 
1012 & 2403 & 44.8 & 7 & 126.7 &&& 17.9 / 36.4 & \p{}7.2 / 10.6 \\ 
\end{tabular}
\end{table}

\paragraph{Focusing on Satisfiable Problems. }

Recall that in our experiments 287 satisfiable problems
were solved by at least one strategy. Table~\ref{tab:sat} (left)
shows the performance of the best 5 \emph{complete} selections on these problems.\footnote{
The table has been shortened due to space restrictions. However, the \#unique and u-score
indicators still take into account all the other complete selections. } 
The first two places are taken by a selection from \E{} and \Spass{}
while the lookahead selection is third. The differences
between the three places are, however, only by one problem. 
Moreover, in Table~\ref{tab:sat} (right) we can see
that when AVATAR was turned off in a separate experiment
lookahead selection came first.


\begin{table}[tb]
\caption{Performance of the five best complete selection strategies on satisfiable problems. 
}
\label{tab:sat}
\centering
\begin{tabular}{rccrr|rccrr}
\multicolumn{5}{c|}{AVATAR on (total 287)} & \multicolumn{5}{|c}{AVATAR off (total 207)} \\ \hline
selection & \#solved & \%total & \#unique & u-score & selection & \#solved & \%total & \#unique & u-score \\
33 & 248 & 86.4 & 0 & 24.5 & 11 & 195 & 94.2 & 0 & 16.7 \\ 
22 & 247 & 86.0 & 0 & 24.1 & 4 & 191 & 92.2 & 0 & 17.1 \\ 
11 & 246 & 85.7 & 0 & 23.4 & 3 & 190 & 91.7 & 0 & 16.9 \\  
32 & 241 & 83.9 & 1 & 23.8 & 32 & 184 & 88.8 & 0 & 14.7 \\ 
1 & 238 & 82.9 & 0 & 21.6 & 35 & 183 & 88.4 & 0 & 14.6 \\  
\end{tabular}
\end{table}

\section{Impact of Selection on Portfolio Solving}

\begin{table}[t]
\caption{Numbers of problems solved only by a strategy using a particular selection strategy.\label{tab:strategy:scheduling}}
\centering
\begin{tabular}{lcc}
selection & \multicolumn{2}{c}{Problems solved only using this selection} \\
\cline{2-3}
& All & Problems solved only by \Vampire\\
\hline
 11& 151 & 118\\
 1011& 78 & 62\\
 1& 62 & 58 \\
 10& 55 & 41\\
 \hline
 lookahead & 278 & 216\\
 non-lookahead & 502 & 377 \\
 complete &824 & 691\\
 incomplete & 229 & 169\\
\end{tabular}

\end{table}

As mentioned at the beginning of Sect.~\ref{sec:experiments}, \Vampire, like most leading first-order theorem provers, will (when asked) try to use a portfolio of \emph{strategies} to solve a problem. To make an effective portfolio we want a mix of strategies that either solve many problems and or unique problems. One measure of the usefulness of a selection strategy is its impact on the creation of a portfolio mode with respect to the second of these properties i.e. which problems can only be solved using a particular approach.

To find useful strategies for \Vampire\ we have a dedicated cluster
using a semi-guided method to randomly search the space of
strategies. At the time of writing over 786k proofs have been found of
11,354 problems out of 13,770 (unsatisfiable, non unit-equality)
problems taken from TPTP 6.1.0. It took over 160 CPU-years of
computation to collect these data. Table~\ref{tab:strategy:scheduling} gives results for the best four selection strategies and four groups of selection strategies. Numbers are given for all problems and for the subset of problems that were not solved by any other theorem prover at the time TPTP 6.1.0 was released\footnote{See the ProblemAndSolutionStatistics file distributed with TPTP.}. From this we can see that both lookahead and incomplete selections are required to solve many problems unsolvable by other methods. 
Additionally, this shows that having a spread of different selection strategies is useful as they contribute to uniquely solving different problems.


\section{Conclusion}
\label{sec:conclusion}

Selection strategies can have a very large impact on proof search, often making the difference between solving and not solving a problem. Little had been written about how effective selection strategies could be designed and implemented, although most successful implementations of the superposition calculus have relied on them.

We have introduced two new ways of performing literal selection based on the observation that it is good to select those literals that lead to as few children as possible. 
The first approach, \emph{quality selection}, is an easy to implement compositional mechanism for defining literal selection based on qualities of literals that lead to few children. 
We described different selection strategies based on concrete qualities and demonstrated their effectiveness. 
What may be surprising to some is how effective \emph{incomplete} versions of such strategies can be. Experimentally establishing this phenomenon is a large contribution of this work. 
However, our main result is the second approach, the powerful idea of \emph{lookahead selection} based on the observation that if we want to select literals leading to as few children as possible then the best thing to do is just that. Experimental results showed that by using this approach we could solve many problems that could not otherwise be solved by any other selection strategy taken from \Vampire, E or \Spass.

\bibliography{bib}
\bibliographystyle{cs-abbrv++}

\end{document}